\documentclass[sigconf]{acmart}
\usepackage{multirow}
\usepackage{graphicx}
\usepackage{booktabs} % For professional-looking table lines (\toprule, \midrule, \bottomrule)
\usepackage{caption}  % For better caption formatting
\usepackage{breakurl}
\AtBeginDocument{%
  }

\setcopyright{acmlicensed}
\copyrightyear{2018}
\acmYear{2018}
\acmDOI{XXXXXXX.XXXXXXX}

\acmConference[Conference acronym 'XX]{Make sure to enter the correct
  conference title from your rights confirmation emai}{June 03--05,
  2018}{Woodstock, NY}
\acmISBN{978-1-4503-XXXX-X/18/06}

\begin{document}

%%
%% The "title" command has an optional parameter,
%% allowing the author to define a "short title" to be used in page headers.
\title{Annotation Tool and Dataset for Fact-Checking Podcasts}

%%
%% The "author" command and its associated commands are used to define
%% the authors and their affiliations.
%% Of note is the shared affiliation of the first two authors, and the
%% "authornote" and "authornotemark" commands
%% used to denote shared contribution to the research.

\author{Vinay Setty}
\affiliation{%
  \institution{University of Stavanger}
  \city{Stavanger}
  \country{Norway}
  }
\email{vsetty@acm.org}

\author{Adam James Becker}
\affiliation{%
  \institution{University of Stavanger}
  \city{Stavanger}
  \country{Norway}
  }
\email{aj.becker@stud.uis.no}

%%
%% By default, the full list of authors will be used in the page
%% headers. Often, this list is too long, and will overlap
%% other information printed in the page headers. This command allows
%% the author to define a more concise list
%% of authors' names for this purpose.
\renewcommand{\shortauthors}{Setty et al.}

%%
%% The abstract is a short summary of the work to be presented in the
%% article.
\begin{abstract}
Podcasts are a popular medium on the web, featuring diverse and multilingual content that often includes unverified claims. Fact-checking podcasts is a challenging task, requiring transcription, annotation, and claim verification, all while preserving the contextual details of spoken content. Our tool offers a novel approach to tackle these challenges by enabling real-time annotation of podcasts during playback. This unique capability allows users to listen to the podcast and annotate key elements, such as check-worthy claims, claim spans, and contextual errors, simultaneously. By integrating advanced transcription models like OpenAI’s Whisper and leveraging crowdsourced annotations, we create high-quality datasets to fine-tune multilingual transformer models such as XLM-RoBERTa for tasks like claim detection and stance classification. Furthermore, we release the annotated podcast transcripts and sample annotations with preliminary experiments. 
\end{abstract}

%%
%% The code below is generated by the tool at http://dl.acm.org/ccs.cfm.
%% Please copy and paste the code instead of the example below.
%%


%%
%% Keywords. The author(s) should pick words that accurately describe
%% the work being presented. Separate the keywords with commas.
%% A "teaser" image appears between the author and affiliation
%% information and the body of the document, and typically spans the
%% page.
% \begin{teaserfigure}
%   \includegraphics[width=\textwidth]{sampleteaser}
%   \caption{Seattle Mariners at Spring Training, 2010.}
%   \Description{Enjoying the baseball game from the third-base
%   seats. Ichiro Suzuki preparing to bat.}
%   \label{fig:teaser}
% \end{teaserfigure}

%%
%% This command processes the author and affiliation and title
%% information and builds the first part of the formatted document.
\maketitle

\section{Introduction}

Podcasts have grown rapidly in popularity worldwide, offering creators significant freedom due to minimal regulation.\footnote{\url{https://web.archive.org/web/20250126182112/https://www.pewresearch.org/journalism/2023/04/18/podcasts-as-a-source-of-news-and-information/}} While this openness fosters creativity, it also increases the risk of misinformation, posing challenges for accurate transcription, contextual understanding, and multilingual support.\footnote{\url{https://web.archive.org/web/20250116085820/https://www.brookings.edu/articles/the-challenge-of-detecting-misinformation-in-podcasting/}} Existing fact-checking methods, which rely on claim-by-claim verification~\cite{thorne-etal-2018-fever,schlichtkrull2023averitec}, are further hindered by the lack of open podcast datasets. Previous datasets like the Spotify Podcast Dataset~\cite{clifton2020100} are no longer accessible, leaving a significant gap.

To address these challenges, we present an open-source tool for podcast transcription and annotation. The tool integrates audio playback with real-time annotation, enabling annotators to correct transcription errors, resolve ambiguities, and identify claims for fact-checking. This approach ensures efficiency while maintaining the integrity of spoken content. Large Language Models (LLMs) are limited in processing multi-hour transcripts and require significant computational resources, but our tool provides a lightweight, accessible alternative.

Built entirely with open-source components, including Whisper ASR~\cite{radford2023robust} for transcription and F-Coref~\cite{otmazgin2022f} for co-reference resolution, the pipeline supports over 90 languages (excluding co-reference resolution) and offers flexibility for annotators to work on multilingual podcasts. It also collects fine-grained annotations, such as claim types, reasons for fact-checking, and utterances unrelated to claims. Annotators can perform fact-checking within the same tool, streamlining the process. We release the source code of the annotation tool here: \url{https://github.com/factiverse/factcheck-podcasts}.

\begin{figure*}[h!]
    \centering
    \includegraphics[width=0.7\linewidth]{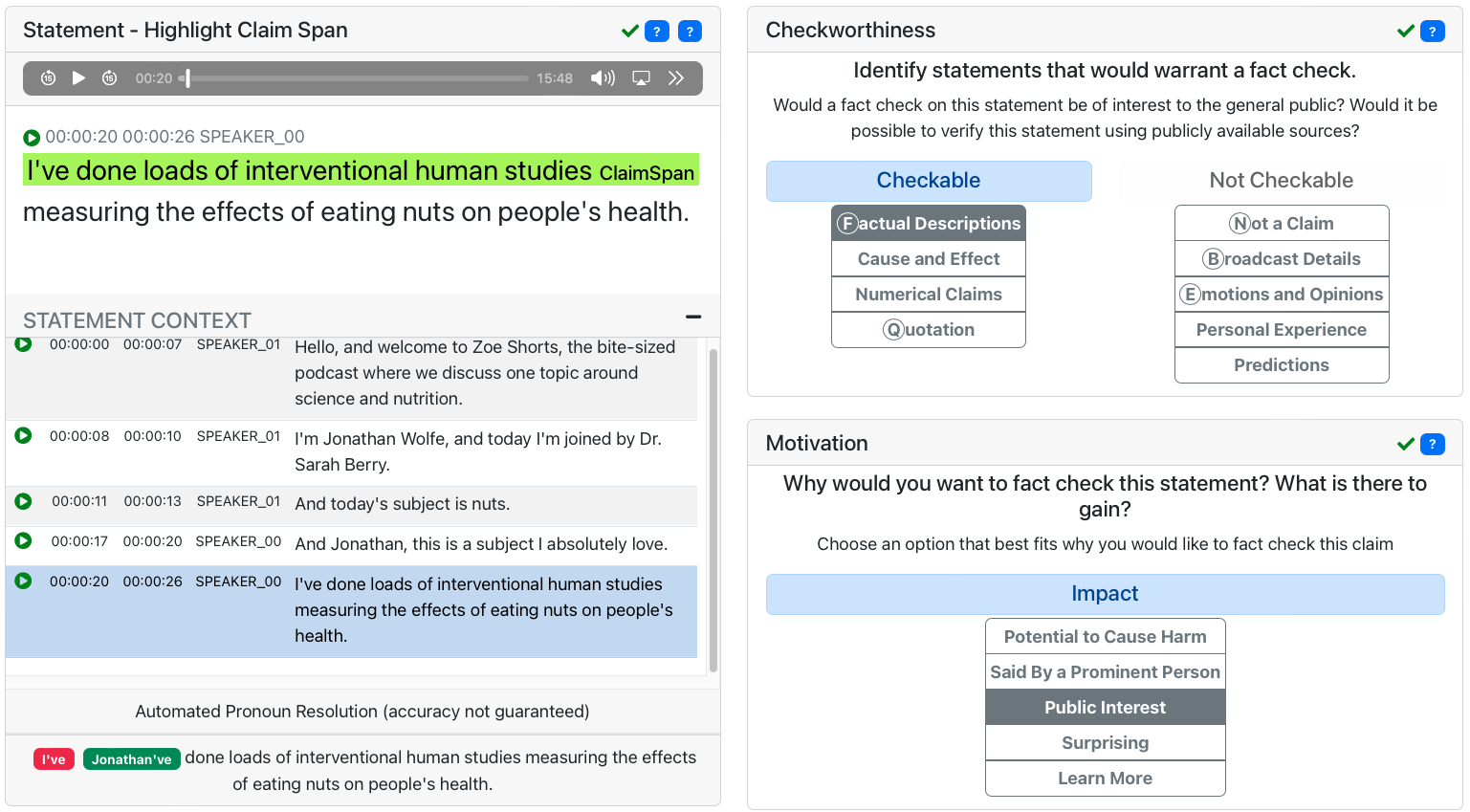}
    \caption{Podcast annotation interface for claim detection. Right hand size shows the options for fine-grained claim annotation.}
    \label{fig:claims}
\end{figure*}

\begin{figure}
    \centering
    \includegraphics[width=0.7\linewidth]{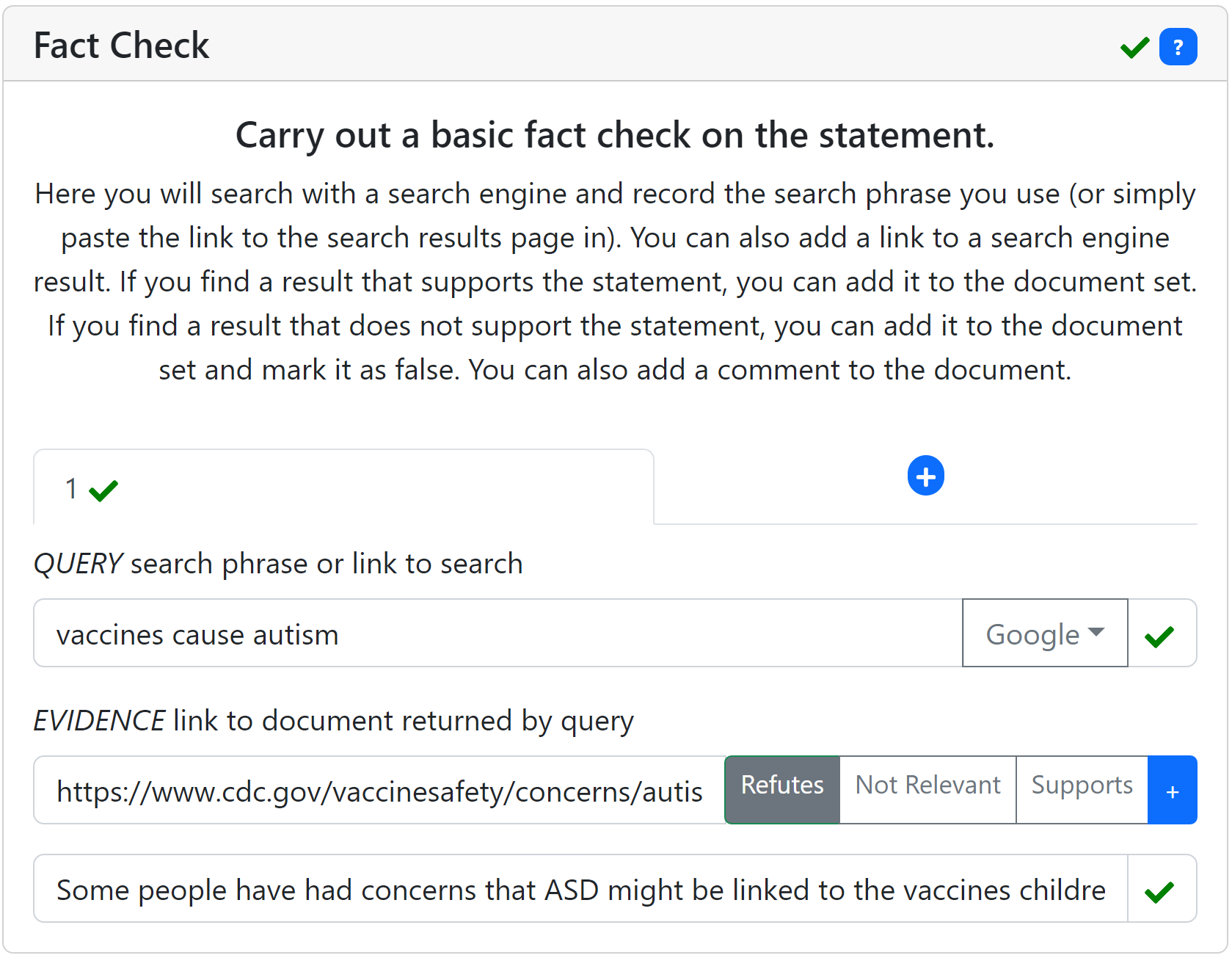}
    \caption{Fact-checking annotation interface.}
    \label{fig:stance}
\end{figure}

We release transcripts for 531 episodes from 38 podcasts in English, Norwegian, and German, alongside an annotated dataset specifically for end-to-end fact-checking. This includes annotations for claim detection, fine-grained claim categorization, and claim verification on selected episodes from 7 podcasts. Additionally, we demonstrate the utility of these annotations by fine-tuning transformer models such as XLM-RoBERTa for multilingual claim detection and stance classification, comparing their performance to LLMs like GPT-4 in few-shot scenarios.

% The annotated data is subsequently utilized to fine-tune transformer models, such as XLM-RoBERTa, facilitating effective multilingual claim detection and stance classification. To promote further research in podcast fact-checking and related fields, we have publicly released the annotated podcast transcripts, encompassing 38 podcasts with 531 episodes in English, Norwegian, and German. Our tool supports over 90 languages (excluding co-reference resolution), simplifying data annotation for a broader range of podcasts across various languages.

\section{Approach}

This section describes the web application for annotating and fact-checking podcasts and the data annotation process.

\subsection{Data Population}

We chose podcasts from two popular categories: \textit{News \& Politics} and \textit{Health \& Wellness}, based on their popularity and relevance. We downloaded podcasts and their metadata from RSS feeds of the original podcast creators. Then we transcribed the podcasts using OpenAI’s Whisper ASR model (with open weights).\footnote{\url{https://github.com/openai/whisper/blob/main/model-card.md}} This model provided outputs that included word-level timestamps, confidence probabilities, and punctuation marks. The outputs were crucial for ensuring accurate alignment between the transcription and the original audio. We enhanced transcriptions by performing diarization using Pyannote~\cite{Plaquet23}, which assigned speaker labels to segments of the audio. These diarization outputs were stored as metadata linked to each transcription.

To facilitate downstream NLP tasks, the transcriptions were segmented into sentences using SpaCy, which split the transcribed text based on punctuation marks. Co-reference resolution was performed using the F-coref package~\cite{otmazgin2022f}. This step ensured that pronouns and proper nouns were linked to the correct entities, creating self-contained utterances that were ready for fact-checking. A sliding window approach was used to process batches of 50 utterances at a time, providing the necessary context for the co-reference resolution model.

\subsection{Crowdsourcing and Annotation}

The web application included an interface for browsing and listening to podcasts, along with a dedicated annotation interface for collecting human annotations on transcriptions (see Figure \ref{fig:claims}). This interface supported tasks such as classifying utterances for check-worthiness, identifying specific claim spans, and correcting transcription errors. Using the interface depicted in Figure \ref{fig:stance}, annotators performed fact-checking by generating search queries, evaluating retrieved documents, and assigning verdicts to claims. Fact-checking was carried out exclusively on claims identified as check-worthy.

We recruited crowdworkers through Prolific, a widely-used crowdsourcing platform, applying pre-screening filters to ensure participants met criteria like language proficiency and educational background. Tasks were broken down into manageable microtasks, typically taking less than 30 minutes to complete. A podcast browser interface provided real-time progress visualization and submission storage, allowing annotators to select podcasts of their choice, reducing annotation fatigue. In the future, we plan to explore more user-friendly annotation methods to further minimize user effort, such as leveraging head-nodding gestures with devices like Apple AirPods.

Each annotation task was assigned to three annotators, with unanimous labels retained to account for subjectivity. Annotations were stored hierarchically in a relational database, linking each annotation to its corresponding utterance and associating it with the relevant podcast episode. This structured storage facilitated flexible querying, enabling the creation of datasets tailored to various NLP applications.

The annotations followed options:

\subsection*{Checkable Claims}
\begin{enumerate}
    \item \textbf{Factual Descriptions:} Verifiable claims about people, places, events, or actions.  
    \textit{Example:} ``She won the London Marathon last year.''
    
    \item \textbf{Cause and Effect:} Claims linking one event to another.  
    \textit{Example:} ``Smoking causes cancer.''
    
    \item \textbf{Numerical Claims:} Claims involving statistics or numerical analysis.  
    \textit{Example:} ``80\% of people are unhappy with the government.''
    
    \item \textbf{Quotations:} Verifiable quotes from notable entities.  
    \textit{Example:} ``President Roosevelt said, `Ich bin ein Berliner.'''
\end{enumerate}

\subsection*{Not Checkable Claims}
\begin{enumerate}
    \item \textbf{Non-Factual Statements:} Phrases without factual assertions.  
    \textit{Example:} ``How are you?''
    
    \item \textbf{Broadcast Details:} Introductions or program descriptions.  
    \textit{Example:} ``Welcome to the show, I'm John Smith.''
    
    \item \textbf{Emotions and Opinions:} Non-verifiable feelings or opinions.  
    \textit{Example:} ``I love how the tulips look in spring.''
    
    \item \textbf{Personal Experience:} Statements about private experiences.  
    \textit{Example:} ``I passed four empty buses yesterday.''
    
    \item \textbf{Predictions:} Future events or plans that cannot be confirmed.  
    \textit{Example:} ``Elon Musk will visit Mars.''
\end{enumerate}

\section*{Motivations for Fact-Checking}
\begin{enumerate}
    \item \textbf{Potential to Cause Harm:} This statement could be harmful if it turns out to be false.
    \item \textbf{Said by a Prominent Person:} I want to verify whether this prominent person actually made this statement.
    \item \textbf{Public Interest:} Fact-checking this claim is important for the public's awareness and benefit.
    \item \textbf{Surprising:} This claim is unexpected, shocking, or difficult to believe without verification.
    \item \textbf{Learn More:} Fact-checking this statement would help me gain a deeper understanding of the topic.
\end{enumerate}

For more details on the annotation process see \cite{becker2023automated}.

\subsection{Application of Annotations}

We analyze the collected annotations and train models for claim detection and stance detection. Check-worthiness classifications and claim spans help identify high-priority claims, while search queries and document evaluations improve evidence retrieval for downstream fact-checking models.

We fine-tune the XLM-Roberta-Large\footnote{\url{https://huggingface.co/FacebookAI/xlm-roberta-large}} model for claim detection and stance detection tasks (NLI). Claim detection is framed as a binary classification task, distinguishing between check-worthy and not check-worthy claims. Similarly, stance detection is simplified to a binary classification task, categorizing claims as either Supports or Refutes.

\section{Dataset Analysis}

In this section, we will analyze the  annotations we conducted on 7 podcast episodes. We plan larger-scale annotations in the future.

\begin{table}[t!!!]
\centering
\caption{Check-worthy and Not Check-worthy Counts by Topic}
\label{tab:Check-worthy_counts}
\begin{tabular}{lrr}
\hline
\textbf{Topic}             & \textbf{Check-worthy} & \textbf{Not Check-worthy} \\ \hline
News and Politics          & 223                  & 1191                     \\ 
Health and Wellness        & 77                   & 469                      \\ 
\midrule
\textbf{Total}             & \textbf{300}         & \textbf{1660}            \\ \hline
\end{tabular}
\end{table}

\subsection{Check-worthy Claim Distribution}

Table \ref{tab:Check-worthy_counts} presents the distribution of utterances identified as \textit{Check-worthy} and \textit{Not Check-worthy} across two main topics: \textit{News and Politics} and \textit{Health and Wellness}. Interestingly, the majority of utterances belong to the \textit{News and Politics} category, where only 223 out of 1414 utterances (approximately 16\%) were deemed Check-worthy.  On the other hand, in \textit{Health and Wellness}, while the total number of utterances is lower (546), 77 utterances (approximately 14\%) were labeled as Check-worthy. This indicates the Check-worthy claims are roughly same proportion across topics.

\begin{table*}[h!!!]
\centering
\caption{Per-Class and Weighted F1-Scores for Claim Detection and Stance Detection Tasks}
\label{tab:eval}
\begin{tabular}{lccc|ccc}
\toprule
\multirow{2}{*}{\textbf{Model}} & \multicolumn{3}{c|}{\textbf{Claim Detection (F1-Score)}} & \multicolumn{3}{c}{\textbf{Stance Detection (F1-Score)}} \\ 
\cline{2-7}
                                & \textbf{False} & \textbf{True} & \textbf{Weighted}         & \textbf{Refutes} & \textbf{Supports} & \textbf{Weighted} \\ 
\midrule
XLM-Roberta-Large (fine-tuned)  & 0.91           & 0.45          & 0.85                      & 0.44             & 0.79              & 0.67              \\ 
GPT-4                           & 0.91           & 0.57          & 0.86                      & 0.53             & 0.56              & 0.55              \\ 
\bottomrule
\end{tabular}
\end{table*}

\subsection{Claim Types}

Numerical Claims dominate with 188 (62.6\%)  of instances, highlighting a focus on statistics and quantitative data. 83 Factual Descriptions (27.6\%) involve verifiable statements about people, events, or objects. Cause and Effect claims account for 19 (6.3\%) are less frequent, and Quotations are rare, with only 10 occurrences, indicating infrequent attribution to specific individuals or entities.

The primary reason statements are not checkable is that the majority (326 utterances) are non-factual, including greetings, questions, or fillers. Emotions and Opinions rank second with 402 utterances, followed by Personal Experiences at 328 utterances. Broadcast Details, such as announcements or advertisements, come next with 306 utterances. Predictions, the smallest category with 60 utterances, involve future events that cannot be verified at present.

\subsection{Motivation to fact-check  claims}

The main motivations for fact-checking are a desire to learn more (127 instances) and public interest (77), reflecting curiosity and the societal importance of verifying claims. Concerns about potential harm (54) also drive fact-checking to prevent misinformation. Interestingly, fewer claims are fact-checked for being surprising facts (24) or said by prominent individuals (18), suggesting that practical and informational needs outweigh sensational or high-profile statements.

\section{Experimental Evaluation}

This section shows the utility of the annotated data for claim detection and stance detection (claim verification) tasks.

\begin{table}[t!!]
    \centering
    \caption{Dataset distribution.}
    \label{tab:dataset}
    \begin{tabular}{l|rrr|rr}
    \hline
      \textbf{Split}   & \textbf{Check-} & \textbf{Not Check-} & \textbf{Total} & \textbf{True} & \textbf{False}   \\ \hline
       Train           & 219                 & 1185         & 1,404   & 323           & 237                     \\
       Dev             & 57                  & 323            & 380   & 86            & 29                       \\
       Test            & 24                  & 152            & 176   & 18            & 32                       \\ \midrule
       \textbf{Total} & \textbf{300} & \textbf{1660}  & \textbf{1960}& \textbf{427} & \textbf{298}\\ \hline
    \end{tabular}
\end{table}

See Table~\ref{tab:dataset} for the dataset used for experiments. We compare the fine-tuned XLM-Roberta-Large and GPT-4 (version gpt-4-0613 deployed on Azure). For the prompt used for GPT-4 refer to \cite{Setty:SIGIR:2024a}.

\subsection{Fact-Checking Results}

From Table \ref{tab:eval}, XLM-Roberta-Large  and GPT-4 both perform well in claim detection, with GPT-4 better at identifying true claims (F1-score of 0.57 vs. 0.45). For stance detection, XLM-Roberta excels, especially in supporting claims (F1-score of 0.79 vs. GPT-4’s 0.56). Overall, GPT-4 is slightly stronger in claim detection, while XLM-Roberta is better in stance detection. This illustrates that smaller LMs are competitive for fact-checking tasks.

\subsection{Transcription Performance}

\begin{table}[h!!]
\centering
\caption{Whisper ASR Performance in terms of error rates (ER)}
\label{tab:asr}
\begin{tabular}{lcccc}
\toprule
       Model Size &  Word ER &  Match ER &  Character ER \\
\midrule
            Small &                   0.25 &                    0.22 &                        0.12  \\
           Medium &                   0.20 &                    0.17 &                        0.10  \\
Medium (Prompted) &                   0.18 &                    0.16 &                        0.09  \\
            Large &                   0.15 &                    0.13 &                        0.07 \\
\bottomrule
\end{tabular}
\end{table}

Table \ref{tab:asr} shows the performance of different sizes of the Whisper ASR model, measured by error rates in word, match, and character. Word Error Rate (WER)  measures the proportion of words incorrectly recognized. Lower WER indicates better performance. Whisper’s error rates improve with model size, with the Small model at 0.25, Medium at 0.20, Medium (Prompted) at 0.18, and Large at 0.15. Match Error Rate measures the order of words and phrases, and Character Error Rate measures the  proportion of individual character errors. Similar pattern is observed in MER (Match Error Rate) and CER (Character Error Rate).

\section{Related Work}

The growing popularity of podcasts has driven research into automated transcription, claim detection, and fact-checking. Early contributions, such as the Spotify Podcast Dataset~\cite{clifton2020100}, provided valuable large-scale spoken document corpora for natural language processing (NLP). Similarly, datasets for podcast summarization, like those proposed by Manakul et al.~\cite{manakul2022podcast}, have advanced understanding in this domain. However, access to these datasets has become increasingly restricted, creating significant gaps in resources for developing robust podcast fact-checking tools.

While datasets for detecting check-worthy claims in political debates exist~\cite{ivanov2024detecting,hassan2017claimbuster}, they lack fine-grained annotations for claim types, motivations for fact-checking, and associated claim verification data, limiting their utility in fact-checking scenarios.

General purpose NLP annotation tools like Prodigy\footnote{\url{https://prodi.gy}} exist. There are also text editors for fact-checking~\cite{setty2024factcheck}, and systems for live fact-checking of audio streams~\cite{setty2024livefc} address various aspects of fact-checking. However, these tools are not designed for data annotation. Currently there are no tools specifically designed for podcast annotation that support simultaneous audio playback and transcription annotation. Additionally, there is a lack of fine-grained analysis of claim types and their motivations for fact-checking. Also, an end-to-end dataset for fact-checking podcasts is currently unavailable.

\section{Conclusion}

This paper presents a novel open-source pipeline for podcast transcription, annotation, and end-to-end fact-checking, addressing critical challenges in long-form audio content. By integrating real-time audio playback with interactive annotation, our tool streamlines the process of correcting transcription errors, identifying check-worthy claims, and resolving contextual ambiguities.

\section*{Acknowledgements}
This research is partially funded by SFI MediaFutures partners and the Research Council of Norway (grant number 309339).
%%
%% The next two lines define the bibliography style to be used, and
%% the bibliography file.
\bibliographystyle{ACM-Reference-Format}
\bibliography{sample-base}

%%% -*-BibTeX-*-
%%% Do NOT edit. File created by BibTeX with style
%%% ACM-Reference-Format-Journals [18-Jan-2012].

\begin{thebibliography}{13}

%%% ====================================================================
%%% NOTE TO THE USER: you can override these defaults by providing
%%% customized versions of any of these macros before the \bibliography
%%% command.  Each of them MUST provide its own final punctuation,
%%% except for \shownote{}, \showDOI{}, and \showURL{}.  The latter two
%%% do not use final punctuation, in order to avoid confusing it with
%%% the Web address.
%%%
%%% To suppress output of a particular field, define its macro to expand
%%% to an empty string, or better, \unskip, like this:
%%%
%%% \newcommand{\showDOI}[1]{\unskip}   % LaTeX syntax
%%%
%%% \def \showDOI #1{\unskip}           % plain TeX syntax
%%%
%%% ====================================================================

\ifx \showCODEN    \undefined \def \showCODEN     #1{\unskip}     \fi
\ifx \showDOI      \undefined \def \showDOI       #1{#1}\fi
\ifx \showISBNx    \undefined \def \showISBNx     #1{\unskip}     \fi
\ifx \showISBNxiii \undefined \def \showISBNxiii  #1{\unskip}     \fi
\ifx \showISSN     \undefined \def \showISSN      #1{\unskip}     \fi
\ifx \showLCCN     \undefined \def \showLCCN      #1{\unskip}     \fi
\ifx \shownote     \undefined \def \shownote      #1{#1}          \fi
\ifx \showarticletitle \undefined \def \showarticletitle #1{#1}   \fi
\ifx \showURL      \undefined \def \showURL       {\relax}        \fi
% The following commands are used for tagged output and should be
% invisible to TeX
\providecommand\bibfield[2]{#2}
\providecommand\bibinfo[2]{#2}
\providecommand\natexlab[1]{#1}
\providecommand\showeprint[2][]{arXiv:#2}

\bibitem[Becker(2023)]%
        {becker2023automated}
\bibfield{author}{\bibinfo{person}{Adam~James Becker}.} \bibinfo{year}{2023}\natexlab{}.
\newblock \emph{\bibinfo{title}{Automated Fact-Checking of Podcasts}}.
\newblock \bibinfo{thesistype}{Master's\ thesis}. \bibinfo{school}{University of Stavanger}.
\newblock
\urldef\tempurl%
\url{https://hdl.handle.net/11250/3088365}
\showURL{%
\tempurl}


\bibitem[Clifton et~al\mbox{.}(2020)]%
        {clifton2020100}
\bibfield{author}{\bibinfo{person}{Ann Clifton}, \bibinfo{person}{Sravana Reddy}, \bibinfo{person}{Yongze Yu}, \bibinfo{person}{Aasish Pappu}, \bibinfo{person}{Rezvaneh Rezapour}, \bibinfo{person}{Hamed Bonab}, \bibinfo{person}{Maria Eskevich}, \bibinfo{person}{Gareth Jones}, \bibinfo{person}{Jussi Karlgren}, \bibinfo{person}{Ben Carterette}, {et~al\mbox{.}}} \bibinfo{year}{2020}\natexlab{}.
\newblock \showarticletitle{100,000 podcasts: A spoken English document corpus}. In \bibinfo{booktitle}{\emph{Proceedings of the 28th International Conference on Computational Linguistics}}. \bibinfo{pages}{5903--5917}.
\newblock


\bibitem[Hassan et~al\mbox{.}(2017)]%
        {hassan2017claimbuster}
\bibfield{author}{\bibinfo{person}{Naeemul Hassan}, \bibinfo{person}{Gensheng Zhang}, \bibinfo{person}{Fatma Arslan}, \bibinfo{person}{Josue Caraballo}, \bibinfo{person}{Damian Jimenez}, \bibinfo{person}{Siddhant Gawsane}, \bibinfo{person}{Shohedul Hasan}, \bibinfo{person}{Minumol Joseph}, \bibinfo{person}{Aaditya Kulkarni}, \bibinfo{person}{Anil~Kumar Nayak}, {et~al\mbox{.}}} \bibinfo{year}{2017}\natexlab{}.
\newblock \showarticletitle{Claimbuster: The first-ever end-to-end fact-checking system}.
\newblock \bibinfo{journal}{\emph{Proceedings of the VLDB Endowment}} \bibinfo{volume}{10}, \bibinfo{number}{12} (\bibinfo{year}{2017}), \bibinfo{pages}{1945--1948}.
\newblock


\bibitem[Ivanov et~al\mbox{.}(2024)]%
        {ivanov2024detecting}
\bibfield{author}{\bibinfo{person}{Petar Ivanov}, \bibinfo{person}{Ivan Koychev}, \bibinfo{person}{Momchil Hardalov}, {and} \bibinfo{person}{Preslav Nakov}.} \bibinfo{year}{2024}\natexlab{}.
\newblock \showarticletitle{Detecting Check-Worthy Claims in Political Debates, Speeches, and Interviews Using Audio Data}. In \bibinfo{booktitle}{\emph{ICASSP 2024-2024 IEEE International Conference on Acoustics, Speech and Signal Processing (ICASSP)}}. IEEE, \bibinfo{pages}{12011--12015}.
\newblock


\bibitem[Manakul and Gales(2022)]%
        {manakul2022podcast}
\bibfield{author}{\bibinfo{person}{Potsawee Manakul} {and} \bibinfo{person}{Mark~JF Gales}.} \bibinfo{year}{2022}\natexlab{}.
\newblock \showarticletitle{Podcast Summary Assessment: A resource for evaluating summary assessment methods}.
\newblock \bibinfo{journal}{\emph{arXiv preprint arXiv:2208.13265}} (\bibinfo{year}{2022}).
\newblock


\bibitem[Otmazgin et~al\mbox{.}(2022)]%
        {otmazgin2022f}
\bibfield{author}{\bibinfo{person}{Shon Otmazgin}, \bibinfo{person}{Arie Cattan}, {and} \bibinfo{person}{Yoav Goldberg}.} \bibinfo{year}{2022}\natexlab{}.
\newblock \showarticletitle{F-coref: Fast, accurate and easy to use coreference resolution}.
\newblock \bibinfo{journal}{\emph{arXiv preprint arXiv:2209.04280}} (\bibinfo{year}{2022}).
\newblock


\bibitem[Plaquet and Bredin(2023)]%
        {Plaquet23}
\bibfield{author}{\bibinfo{person}{Alexis Plaquet} {and} \bibinfo{person}{Hervé Bredin}.} \bibinfo{year}{2023}\natexlab{}.
\newblock \showarticletitle{{Powerset multi-class cross entropy loss for neural speaker diarization}}. In \bibinfo{booktitle}{\emph{Proc. INTERSPEECH 2023}}.
\newblock


\bibitem[Radford et~al\mbox{.}(2023)]%
        {radford2023robust}
\bibfield{author}{\bibinfo{person}{Alec Radford}, \bibinfo{person}{Jong~Wook Kim}, \bibinfo{person}{Tao Xu}, \bibinfo{person}{Greg Brockman}, \bibinfo{person}{Christine McLeavey}, {and} \bibinfo{person}{Ilya Sutskever}.} \bibinfo{year}{2023}\natexlab{}.
\newblock \showarticletitle{Robust speech recognition via large-scale weak supervision}. In \bibinfo{booktitle}{\emph{International conference on machine learning}}. PMLR, \bibinfo{pages}{28492--28518}.
\newblock


\bibitem[Schlichtkrull et~al\mbox{.}(2023)]%
        {schlichtkrull2023averitec}
\bibfield{author}{\bibinfo{person}{Michael Schlichtkrull}, \bibinfo{person}{Zhijiang Guo}, {and} \bibinfo{person}{Andreas Vlachos}.} \bibinfo{year}{2023}\natexlab{}.
\newblock \showarticletitle{AVeriTeC: A dataset for real-world claim verification with evidence from the web}.
\newblock \bibinfo{journal}{\emph{arXiv preprint arXiv:2305.13117}} (\bibinfo{year}{2023}).
\newblock


\bibitem[Setty(2024a)]%
        {setty2024factcheck}
\bibfield{author}{\bibinfo{person}{Vinay Setty}.} \bibinfo{year}{2024}\natexlab{a}.
\newblock \showarticletitle{Factcheck editor: Multilingual text editor with end-to-end fact-checking}. In \bibinfo{booktitle}{\emph{Proceedings of the 47th International ACM SIGIR Conference on Research and Development in Information Retrieval}}. \bibinfo{pages}{2744--2748}.
\newblock


\bibitem[Setty(2024b)]%
        {Setty:SIGIR:2024a}
\bibfield{author}{\bibinfo{person}{Vinay Setty}.} \bibinfo{year}{2024}\natexlab{b}.
\newblock \showarticletitle{Surprising Efficacy of Fine-Tuned Transformers for Fact-Checking over Larger Language Models}. In \bibinfo{booktitle}{\emph{Proceedings of the 47th International ACM SIGIR Conference on Research and Development in Information Retrieval}} \emph{(\bibinfo{series}{SIGIR '24})}. \bibinfo{pages}{2842–2846}.
\newblock


\bibitem[Thorne et~al\mbox{.}(2018)]%
        {thorne-etal-2018-fever}
\bibfield{author}{\bibinfo{person}{James Thorne}, \bibinfo{person}{Andreas Vlachos}, \bibinfo{person}{Christos Christodoulopoulos}, {and} \bibinfo{person}{Arpit Mittal}.} \bibinfo{year}{2018}\natexlab{}.
\newblock \showarticletitle{{FEVER}: a Large-scale Dataset for Fact Extraction and {VER}ification}. In \bibinfo{booktitle}{\emph{Proceedings of the 2018 Conference of the North {A}merican Chapter of the Association for Computational Linguistics: Human Language Technologies, Volume 1 (Long Papers)}}. \bibinfo{pages}{809--819}.
\newblock


\bibitem[V and Setty(2024)]%
        {setty2024livefc}
\bibfield{author}{\bibinfo{person}{Venktesh V} {and} \bibinfo{person}{Vinay Setty}.} \bibinfo{year}{2024}\natexlab{}.
\newblock \showarticletitle{LiveFC: A System for Live Fact-Checking of Audio Streams}.
\newblock  (\bibinfo{year}{2024}).
\newblock
\showeprint[arxiv]{2408.07448}~[cs.CL]


\end{thebibliography}

%%
%% If your work has an appendix, this is the place to put it.

\end{document}